\documentclass[preprint,12pt]{elsarticle}

\usepackage{amssymb}

\usepackage{booktabs}
\usepackage{textcomp}
\usepackage{siunitx}
\usepackage[acronym]{glossaries}
\usepackage{subcaption}
\usepackage{url}
\usepackage[breaklinks=true]{hyperref}
\usepackage{cleveref}
\glsdisablehyper
\newacronym{ml}{ML}{Machine Learning}
\newacronym{nn}{NN}{Neural Network}
\newacronym{fpga}{FPGA}{Field Programmable Gate Array}
\newacronym{asic}{ASIC}{Application Specific Integrated Circuit}
\newacronym{vww}{VWW}{Visual Wake Words}
\newacronym{miap}{MIAP}{More Inclusive Annotations for People}
\newacronym{url}{URL}{Uniform Resource Locator}
\newacronym{coco}{COCO}{Common Objects in Context}
\newacronym{tfds}{TFDS}{TensorFlow Datasets}
\newacronym{bbox}{bbox}{Bounding Box}
\newacronym{gpu}{GPU}{Graphics Processing Unit}
\newacronym{pdm}{PdM}{Predictive Maintenance}
\newacronym{nas}{NAS}{Neural Architecture Search}
\newacronym{dnas}{DNAS}{Differentiable Neural Architecture Search}
\newacronym{mfcc}{MFCC}{Mel-Frequency Cepstral Coefficients}
\newacronym{nre}{NRE}{Non-Recurrent Engineering}
\newacronym{rul}{RUL}{Remaining Useful Life}
\newacronym{cnn}{CNN}{Convolutional Neural Network}
\newacronym{tf}{TF}{TensorFlow}
\newacronym{tflm}{TFLM}{TensorFlow Lite Micro}
\newacronym{knn}{KNN}{k-Nearest Neighbor}
\newacronym{kd}{KD}{Knowledge Distillation}
\newacronym{roc}{ROC}{Receiver Operating Characteristic}
\newacronym{auc}{AUC}{Area Under Curve}
\newacronym{stft}{STFT}{Short Time Fourier Transform}
\newacronym{svm}{SVM}{Support Vector Machine}
\newacronym{ann}{ANN}{Artificial Neural Network}
\newacronym{rnn}{RNN}{Recurrent Neural Network}
\newacronym{tfl}{TFL}{TensorFlow Lite}
\newacronym{ell}{ELL}{Embedded Learning Library}
\newacronym{aifes}{AIfES}{Artificial Intelligence for Embedded Systems}
\newacronym{fann}{FANN}{Fast Artificial Neural Network}
\newacronym{pulp}{PULP}{Parallel Ultra Low Power}
\newacronym{onnx}{ONNX}{Open Neural Network Exchange}
\newacronym{cntk}{CNTK}{Microsoft Cognitive Toolkit}
\newacronym{ip}{IP}{Intellectual Property}
\newacronym{gpio}{GPIO}{General Purpose Input/Output}
\newacronym{relu}{ReLU}{Rectified Linear Unit}
\newacronym{soc}{SoC}{System on a Chip}
\newacronym{xai}{XAI}{Explainable AI}
\newacronym{cpu}{CPU}{Central Processing Unit}
\newacronym{dsp}{DSP}{Digital Signal Processor}
\newacronym{simd}{SIMD}{Single Instruction Multiple Data}
\newacronym{mac}{MAC}{Multiply-Accumulate}
\newacronym{ic}{IC}{Integrated Circuit}
\newacronym{sram}{SRAM}{Static RAM}
\newacronym{dram}{DRAM}{Dynamic RAM}
\newacronym{mlp}{MLP}{Multi-Layer Perceptron}
\newacronym{automl}{AutoML}{Automated Machine Learning}
\newacronym{un}{UN}{United Nations}
\newacronym{ble}{BLE}{Bluetooth Low Energy}
\newacronym{it}{IT}{Information Technology}
\newacronym{ai}{AI}{Artificial Intelligence}
\newacronym{mcu}{MCU}{Microcontroller Unit}
\newacronym{rgb}{RBG}{Red, Green, and Blue}
\newacronym{iot}{IoT}{Internet of Things}

\begin{document}

\begin{frontmatter}



\title{Fast Data Aware Neural Architecture Search via 
Supernet Accelerated Evaluation}

\author[inst1]{Emil Njor \fnref{email}}
\fntext[email]{\emph{Email Address: }\texttt{emjn@dtu.dk} (Emil Njor)}

\affiliation[inst1]{organization={Technical University of Denmark},
            addressline={Anker Engelunds Vej 101}, 
            city={Kongens Lyngby},
            postcode={2800}, 
            country={Denmark}}

\affiliation[inst2]{organization={Microsoft},
            addressline={1 Memorial Dr}, 
            city={Cambridge, MA},
            postcode={02142}, 
            country={US}}

\author[inst2]{Colby Banbury}

\author[inst1]{Xenofon Fafoutis}

\begin{abstract}
Tiny machine learning (TinyML) promises to revolutionize fields such as healthcare, environmental monitoring, and industrial maintenance by running machine learning models on low-power embedded systems.
However, the complex optimizations required for successful TinyML deployment continue to impede its widespread adoption.

A promising route to simplifying TinyML is through automatic machine learning (AutoML), which can distill elaborate optimization workflows into accessible key decisions.
Notably, Hardware Aware Neural Architecture Searches --- where a computer searches for an optimal TinyML model based on predictive performance and hardware metrics --- have gained significant traction, producing some of today's most widely used TinyML models.

Nevertheless, limiting optimization solely to neural network architectures can prove insufficient.
Because TinyML systems must operate under extremely tight resource constraints, the choice of input data configuration, such as resolution or sampling rate, also profoundly impacts overall system efficiency. 
Achieving truly optimal TinyML systems thus requires jointly tuning both input data and model architecture.

Despite its importance, this ``Data Aware Neural Architecture Search'' remains underexplored. 
To address this gap, we propose a new state-of-the-art Data Aware Neural Architecture Search technique and demonstrate its effectiveness on the novel TinyML ``Wake Vision'' dataset.
Our experiments show that across varying time and hardware constraints, Data Aware Neural Architecture Search consistently discovers superior TinyML systems compared to purely architecture-focused methods, underscoring the critical role of data-aware optimization in advancing TinyML.
\end{abstract}

\begin{graphicalabstract}
\includegraphics[width=\linewidth]{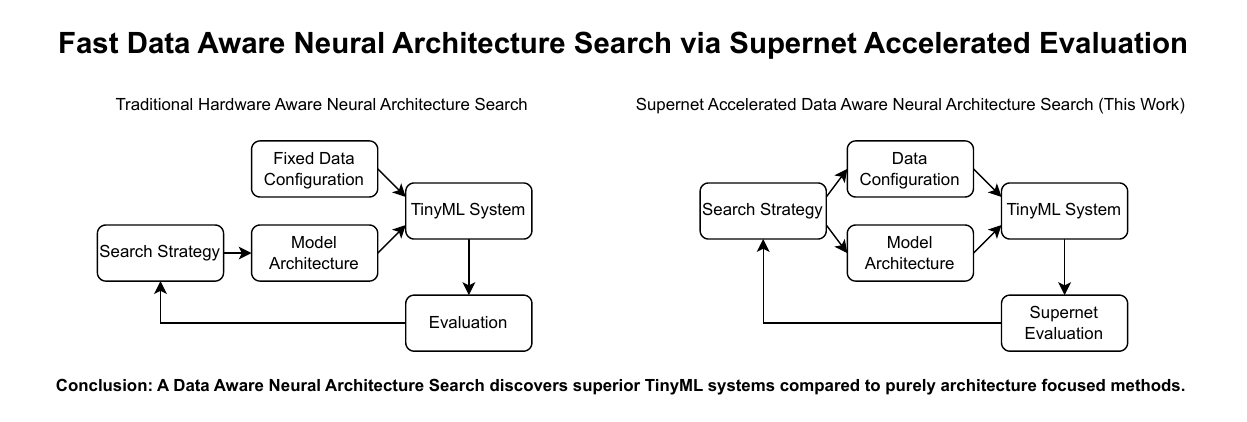}
\end{graphicalabstract}

\begin{highlights}
\item Expands research on combining a Neural Architecture Search with a search for a data configuration called ``Data Aware Neural Architecture Search'' for creating lean TinyML systems.
\item Novel contributions include a new Supernet Based Data Aware Neural Architecture Search algorithm that significantly increases the speed of the search.
\item Applying the new algorithm alongside the larger and more complex Wake Vision dataset shows it consistently discovering superior TinyML systems across varying time and hardware constraints compared to purely architecture-focused methods.
\end{highlights}

\begin{keyword}
TinyML \sep Neural Architecture Search \sep Data Aware Machine Learning \sep Embedded Machine Learning
\MSC 68T10 \sep 68T20 \sep 68T45
\end{keyword}

\end{frontmatter}


\section{Introduction}
\label{sec:introduction}
Today, low-power microcontrollers are ubiquitous, embedded within \gls{iot} systems where they perform specialized tasks such as controlling appliances and monitoring sensors~\cite{saha2022machine}.
By integrating intelligent \gls{ml} systems into these microcontrollers, we can enable them to make intelligent decisions, and adapt to changing conditions, at the edge.
This empowers devices to anticipate needs, simplify interactions, and ultimately improve quality of life.
This vision is at the core of the TinyML field, which has already demonstrated significant impact across diverse applications such as environmental monitoring~\cite{ITU_tinyml_2024}, predictive maintenance~\cite{njor2024holistic}, and healthcare~\cite{tsoukas2021review}.

Successfully deploying \gls{ml} systems --- particularly deep learning systems, which are inherently resource-intensive --- on microcontrollers is challenging.
These low-power devices' strict limitations on compute, power, and memory demand substantial optimizations to produce feasible systems.
Such optimizations necessitate tailoring the hardware platform, inference library, \gls{ml} model, and data pre-processing pipeline to meet application-specific resource constraints --- a process that, with current tools, demands significant \gls{nre} effort.

This challenge is further exacerbated by the heterogeneous landscape of microcontrollers, which consist of a wide variety of models and manufacturers.
While individual microcontrollers are inexpensive, replacing a large fleet of deployed devices --- especially those in remote and hard-to-reach locations --- can be prohibitively costly or impractical.
Consequently, TinyML systems must often operate on heterogeneous hardware platforms, each requiring customized \gls{nre} work to fit varying resource capabilities.

The high \gls{nre} costs associated with deploying TinyML systems create a significant barrier to adoption.
To address this, AutoML techniques have emerged as a promising solution for automating the optimization process. 
In particular, Hardware Aware \gls{nas}~\cite{benmeziane2021comprehensive} is regarded as a promising research direction to scale the process of creating lean TinyML systems.
Alongside scaling the creation of TinyML systems, the speed at which \gls{nas} evaluates \gls{nn} architectures has also proven useful for creating higher-performing architectures than manually designed architectures.
Models such as MCUNet~\cite{lin2020mcunet} and Micronets~\cite{banbury2021micronets}, both designed through Hardware-Aware \gls{nas}, are widely recognized benchmarks for resource-efficient TinyML systems.

While Hardware-Aware \gls{nas} optimizes the resource usage of neural networks, it considers only the model architecture and overlooks the role of input data configurations --- another critical component of TinyML systems.
To bridge this gap, researchers have recently introduced ``Data-Aware \gls{nas}''~\cite{njor2023data}, an approach that extends Hardware-Aware \gls{nas} by simultaneously searching for the optimal input data configuration alongside the model architecture~\cite{njor2023data}.

In this work, we continue this thrust towards Data Aware \gls{nas} by showing that this extension offers two key advantages:

\paragraph{Improved Resource Efficiency} Data Aware \gls{nas} can reduce data granularity, leading to smaller input data, intermediate \gls{nn} representations, and model parameters.
This reduction can minimize resource usage while still enabling the system to meet predictive performance constraints.

\paragraph{Enhanced Predictive Performance} Under strict resource constraints, Data-Aware \gls{nas} can identify optimal trade-offs between data granularity and model complexity, resulting in improved predictive performance compared to traditional Hardware-Aware \gls{nas} approaches.
\newline

Expanding the \gls{nas} search space to include input data configurations increases the search space size manyfold.
While this could have introduced the challenge of maintaining tractable search times, our experiments show that a Data Aware \gls{nas} outperforms a traditional Hardware Aware \gls{nas} across varying time constraints.

In their initial exploration of Data Aware \gls{nas}~\cite{njor2023data}, the authors developed a simple layer-based approach and validated its effectiveness on the ToyADMOS dataset~\cite{koizumi2019toyadmos}.
Although this work successfully demonstrated the Data Aware \gls{nas} concept, the simplicity of the ToyADMOS dataset limits the ability to showcase the full potential of Data Aware \gls{nas} under more realistic and challenging scenarios.

In this paper, we advance the state of Data Aware \gls{nas} by applying it to the significantly more complex Wake Vision dataset~\cite{banbury2024wake}.
Exploring TinyML systems for using this dataset would be infeasible using the prior Data Aware \gls{nas} implementation due to speed limitations. 
Therefore, to overcome this barrier, we integrate state-of-the-art Supernet-based \gls{nas} techniques in a new Data Aware \gls{nas} implementation to ensure tractable search times.
Specifically, we make the following contributions:

\paragraph{Improved Performance Under Resource Constraints} 
We demonstrate that Data-Aware \gls{nas} consistently outperforms traditional Hardware-Aware \gls{nas} in creating TinyML systems that achieve better predictive performance within the same resource constraints.

\paragraph{Efficient Search Times}
We show that Data-Aware \gls{nas} discovers superior TinyML systems compared to purely architecture-focused methods regardless of time constraints.

\paragraph{Reduced Expert Intervention}
Compared to Hardware-Aware \gls{nas}, our approach reduces the expert knowledge required to design and create efficient TinyML systems.

\paragraph{Supernet-based Evaluation}
We contribute an improved Data Aware \gls{nas} based on supernets, significantly accelerating the \gls{nas} evaluation process and enabling the discovery of optimized systems for complex datasets and search spaces.
\newline

The remaining sections of the paper are organized as follows:
\paragraph{Section 2: Related Work} Presents prior scientific works that closely align with the challenges and solutions presented in this work. 
\paragraph{Section 3: Resource Consumption Analysis} Examines the typical resource constraints encountered in TinyML and analyzes the resource interplay between data and TinyML models.
\paragraph{Section 4: Improving Data Aware Neural Architecture Search} Introduces the enhanced Data Aware \gls{nas} implementation enabling the experiments presented in this work.
\paragraph{Section 5: Results} Presents the experimental outcomes of our enhanced Data Aware \gls{nas} approach and discusses the insights drawn from them.
\paragraph{Section 6: Future Work} Identifies promising directions for extending this research.
\paragraph{Section 7: Conclusion} Highlights key findings and contributions of the work.
\paragraph{Section 8: Acknowledgements} Recognizes the funding sources that enabled this research.
\paragraph{Section 9: Declaration of generative AI and AI-assisted technologies in the writing process} Discloses the use of generative \gls{ai} for refining the manuscript's language.

\section{Related Work}
\label{sec:related_work}
The research presented in this paper intersects multiple fields.
First, it is based on the idea of using a computer algorithm to optimize \gls{nn} architectures, commonly known as \gls{nas}.
Second, it considers the deployment of resource-intensive \gls{nn} models on severely resource-constrained embedded devices, a domain known as TinyML.
This has so far been achieved either by handcrafting manual \gls{nn} architectures or via Hardware Aware \glspl{nas}.
Third, it intersects the growing trend of Data-Centric \gls{ai}.

\subsection{Neural Architecture Search}
\Gls{nas} is a widely explored approach for designing high-performance \gls{nn} architectures in traditional computational scales~\cite{zoph2017neural, zoph2018learning, ying2019bench, brock2018smash}.
Early \gls{nas} systems relied on evolutionary algorithms~\cite{real2017large} and reinforcement learning~\cite{zoph2017neural, zoph2018learning} to explore discrete search spaces of potential \gls{nn} architectures.

Recently, supernet-based methods, which train a single large model and extract high-quality subarchitectures from it, have gained popularity as a way to reduce search times~\cite{brock2018smash}.
\Gls{dnas} has also emerged as a promising technique, where a discrete \gls{nn} search space is turned continuous through proxy scalar values, enabling fast searches through gradient-based optimization methods~\cite{liu2018darts}.

\subsection{TinyML Neural Architecture Search}
The application of \gls{nas} has been central in creating models that balance performance and efficiency to run on TinyML hardware since the very beginning of the field.
SpArSe~\cite{fedorov2019sparse}, one of the first works to show that \glspl{cnn} can be deployed on TinyML hardware, uses bayesian-based \gls{nas} to find suitable \gls{cnn} architectures.

More recent works on TinyML \gls{nas} include MCUNet~\cite{lin2020mcunet} and Micronets~\cite{banbury2021micronets}, used to create the MCUNet and Micronet family of models, which are among the most used in TinyML.

The MCUNet \gls{nas} consists of two steps.
During the first step, a search space is created where most models fit tightly into the resource constraints of a TinyML device.
In the second step, a one-shot \gls{nas} is performed on the search space, after which an evolutionary search is used to sample the best sub-network.
The MCUNet \gls{nas} arguably includes data in its search space as the adapted search space for TinyML is created by, among others, reducing the input resolution of the search space~\cite{lin2020mcunet}.

The authors of the Micronets paper use \gls{dnas}~\cite{liu2018darts} based on either a MobileNetv2~\cite{sandler2018mobilenetv2} or a DS-CNN(L)~\cite{zhang2017hello} backbone to design TinyML models for Person Detection, Keyword Spotting, and Anomaly Detection.

Other works on \gls{nas} for TinyML include CNAS, which optimizes \gls{nn} for hardware device operator constraints~\cite{gambella2022cnas} and \textmu{}NAS, which proposes a \gls{nas} with a highly granular search space~\cite{liberis2021munas}.

\subsection{Manually Designed TinyML Architectures}
While \gls{nas} dominates TinyML model development, several works propose manually designed architectures.
For instance, CoopNet is a manually designed model that utilizes heterogeneous quantization and binarization to produce an efficient model architecture~\cite{mocerino2019coopnet}.
Another example is the DS-CNN model, which uses depthwise separable convolution layers to achieve superior performance in keyword spotting applications within TinyML resource constraints~\cite{zhang2017hello}.

\subsection{Data-Centric Artificial Intelligence}
An up-and-coming research area that has traditionally received little attention is Data-Centric \gls{ai}.
Where many research papers propose techniques for optimizing \gls{ml} models, much fewer papers consider ways to improve \gls{ml} through data-centric innovations~\cite{mazumder2023dataperf}.
Techniques such as weak supervision for fast dataset labeling~\cite{ratner2017snorkel}, confident learning for automated label error spotting and correction~\cite{northcutt2021confident}, and active learning where only the most important data samples are labeled~\cite{settles2009active} are key contributions to this domain.

\subsection{Data Aware Neural Architecture Search}
Researchers recently introduced Data Aware \gls{nas}~\cite{njor2023data}, which incorporates data configurations into the search space of a Hardware Aware \gls{nas}.
Using evolutionary algorithms, they jointly optimized data configurations and \gls{cnn} models within a layer-based search space.
Their findings demonstrates that Data Aware \gls{nas} can significantly reduce resource consumption in TinyML systems.

Building on this foundation, this present study conducts a comprehensive comparison between Data Aware- and traditional Hardware-Aware \gls{nas} through experiments on a more complex dataset.
This comparison would not be possible using the prior Data Aware \gls{nas} implementation due to speed limitations.
To overcome this barrier, we contribute a new Data Aware \gls{nas} implementation, built on state-of-the-art \gls{nas} techniques.

\section{Resource Consumption Analysis}
\label{sec:resource_consumption}
Prior work on Data Aware Neural Architecture Search~\cite{njor2023data} did not comprehensively analyze the resource consumption dynamics between sensor data and TinyML models.
To address this gap, this section explores the resource consumption of TinyML systems in depth.
We first characterize the available resources in such systems, then examine how data transfer from sensors impacts resource usage, and finally analyze the resource consumption of running TinyML models. 

\subsection{Available Resources}
The available resources in a TinyML system are primarily decided by the hardware platform on which it is deployed.
Typical resources that we consider are computational capacity, volatile RAM, persistent flash storage, and energy.
These resources are interdependent, with energy consumption often being linked to the usage of other resources.

Consider the Arduino Nano 33 BLE Sense~\cite{arduino-nano}, a popular device in the TinyML community.
It features an ARM Cortex M4 processor supporting \gls{simd} parallel processing for operations like \glspl{mac}, commonly used in \glspl{nn}.
The device provides \SI{256}{\kilo\byte} of \gls{sram} and \SI{1}{\mega\byte} of flash storage.
While there are no official figures for the energy consumption of the device, our measurements show a power draw of around \SI{125}{\milli\watt} during \gls{ml} inference.

\subsection{Sensor Data Resource Consumption}
In a TinyML system, sensor data undergoes several steps: capturing, transferring to the \gls{mcu} memory, pre-processing, and feeding into the TinyML model.
Fast and Efficient data transfer throughout these steps requires fast read and write operations.
Among the memory types typically present in \glspl{mcu}, RAM fits these characteristics best.

Pre-processing sensor data uses both processing and memory resources, requiring data to be read, processed, and written back to memory.
Feeding the processed data to a TinyML model may require further data movement if the model requires input data to be stored in a specific location.

Data can be fed to models using two primary paradigms:
\begin{description}
    \item[Streaming] Data is processed one sample at a time as it is captured.
    Also known as online processing.
    \item[Batched] Data is accumulated in a buffer until enough samples are collected for parallel processing.
    Also known as offline processing.
\end{description}

The batch paradigm consumes more memory as sensor data needs to be aggregated and preserved in memory for longer than with streaming. 
However, batched processing can leverage improved parallel processing and, therefore, improve the inference efficiency of the model.
Furthermore, batching of sensor data is required to predict over the temporal dimension (e.g., audio data), except in some rare instances where recurrent neural networks are used.
Due to the tight memory constraints of TinyML systems, large batches are typically infeasible, and streaming is used more frequently.

\subsection{Estimating TinyML Model Resource Consumption}
Compute, memory, storage, and energy are precious resources for TinyML systems, making it crucial to track and estimate their usage accurately.
Most embedded applications statically allocate memory and storage. 
Therefore, we can accurately predict the consumption of each by the architecture of the deployed \gls{nn}.
For storage (i.e., non-volatile flash) consumption, the primary contributor is the number of parameters in the model and their data type (e.g., int8, fp16, etc.). 
As shown in \cref{eq:flash_consumption}, one can sum the number of parameters in a model, multiply by the size of each parameter, which is determined by the datatype, and factor in some overhead for the rest of the application stack~\cite{banbury2021micronets}.
You can predict the working memory consumption of a model using \cref{eq:ram_consumption}. You take the maximum of the input and output intermediate tensors for any layer in a network, factoring in any residuals or other tensors that need to be preserved in memory.

\begin{equation}
\begin{aligned}
    c_{f} &= \sum{m_{p}t_{s}} \\
    &\text{where} \quad
    c_{f} \quad \text{: Flash Consumption in Bytes,} \\
    m_{p} &\quad \text{: Model Parameter,} \\
    t_{s} &\quad \text{: Size of Parameter Datatype in Bytes.} 
\end{aligned}  
\label{eq:flash_consumption}
\end{equation}

\begin{equation}
\begin{aligned}
    c_{r} &= \max\left(d + l_{1o} \in m, \max_{l_{j} \in m}\left(l_{ji} + l_{jo}\right)\right)t_{s} \\
    &\text{where} \quad
    c_{r} \quad \text{: RAM Consumption in Bytes,} \\ 
    d &\quad \text{: Size of Input Data in Bytes,} \\
    l_{j} &\quad \text{: A layer in a Model,} \\
    m &\quad \text{: The set of all layers in a Model,} \\
    l_{1o} &\quad \text{: Size of First Model Layer Output in Bytes,} \\
    l_{i} &\quad \text{: Size of Layer Input in Bytes,} \\
    l_{o} &\quad \text{: Size of Layer Output in Bytes,} \\
    t_{s} &\quad \text{: Size of Data, Input and Output Datatypes in Bytes.} 
\end{aligned}
\label{eq:ram_consumption}
\end{equation}

Computational resources are not ``consumed'' in the same way, but they determine if a given model is computed within the latency budget of the application. 
To predict latency, you can often use proxy metrics, such as the number of operations, or predictive surrogate models~\cite{benmeziane2021comprehensive}.
To get the best accuracy, one must measure the real latency on the device itself.
The same goes for energy consumption, which, given the relatively simple hardware architecture of MCUs, is inextricably linked to the latency.

\section{Improving Data Aware Neural Architecture Search}\label{sec:improved_dnas}
While prior work demonstrated that a Data Aware \gls{nas} could reduce resource consumption of TinyML systems under predictive performance constraints, it failed to show the trading-off of data and model resources to improve predictive performance~\cite{njor2023data}.

To show this phenomenon in this paper, we must target a less trivial application than previously~\cite{njor2023data}.
A promising application in TinyML is Person Detection, where an image is to be classified according to whether it contains a person~\cite{banbury2021mlperf}.
As this is a vision application, it will work on higher dimensionality data than previously considered in Data Aware \gls{nas}, enabling a deeper exploration of resource-performance trade-offs.

Targeting Person Detection allows us to utilize the new ``Wake Vision'' dataset~\cite{banbury2024wake}, recently introduced to the TinyML domain, promising vast improvements in size and quality over traditional TinyML datasets.

\subsection{A Supernet Based Data Aware Neural Architecture Search}
The increased complexity of Person Detection, coupled with the larger Wake Vision dataset, makes the prior approach of using a layer-based search space and naive evolutionary algorithms too slow for practical use.
To address this, we propose a new supernet-based implementation utilizing a MobileNetv2~\cite{sandler2018mobilenetv2} backbone.

In this approach, a fully sized MobileNetv2 supernet is trained for each input data configuration.
Training is performed lazily, meaning it is initiated only right before the supernet's first use.
During evaluation, the Data Aware \gls{nas} samples a model configuration from the supernet corresponding to its data configuration, which provides a model with pre-trained weights.
This model requires only minimal fine-tuning, enabling faster evaluation~\cite{cai2018proxylessnas}.

We evaluate each TinyML system using the fitness function described in \ref{app:fitness_function_details}, and use a tournament-based genetic algorithm to produce new candidate TinyML systems based on previous ones.

\paragraph{Input Data Configuration Options}
We configure the new Data Aware \gls{nas} implementation to consider two orthogonal ways to configure its data.
First, we allow it to configure the resolution of the image that is passed onto the \gls{nn} model.
Possible resolution options span $R=\{32,64,96,128,160,192,224\}$.
Large input resolutions profoundly impact the RAM consumption of data due to the increased shape of the tensor storing the image.
It also has a significant effect on the size, and therefore RAM consumption, of the internal representations, i.e., activations, of the \gls{nn} model created by the convolutional and depthwise separable convolutional layers, both of which are frequent in a MobileNetV2 model.
The number of parameters of a model is less affected by input resolution changes, as parameters in convolutional and depthwise separable convolutional layers depend on fixed filter sizes, not tensor dimensions.

Secondly, we enable the Data Aware \gls{nas} to adapt whether the image is encoded in monochrome or \gls{rgb} format, i.e., $C=\{\text{monochrome},\text{rbg}\}$.
If a monochrome image encoding is picked, we reduce the width (also called alpha)~\cite{sandler2018mobilenetv2} of the MobileNetv2 supernet by $1/3$.
Consequently, this change affects both the RAM and flash consumption of the final TinyML system.

\begin{figure}
    \centering
    \includegraphics[width=0.8\linewidth]{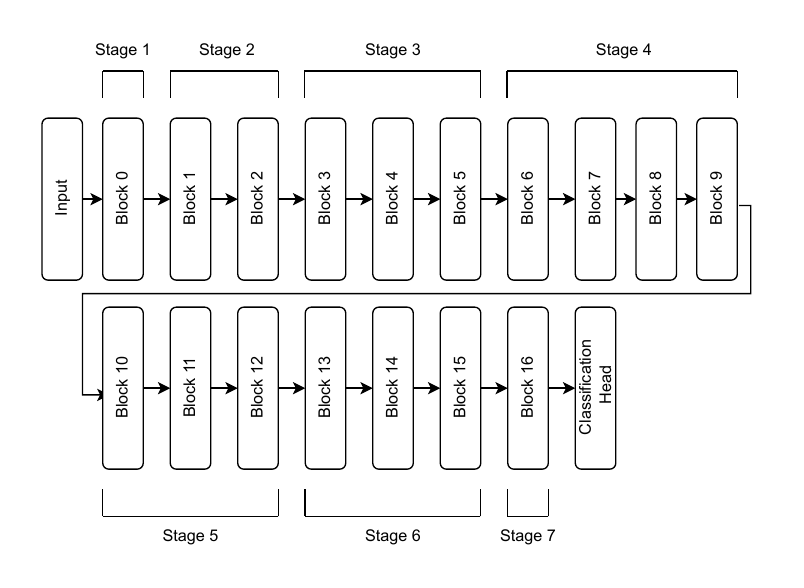}
    \caption{The structure of a MobileNetV2 Model. Each block consists of an inverted bottleneck.}
    \label{fig:mobilenet-figure}
\end{figure}

\paragraph{Model Configuration Options}
We consider two ways to adapt the MobileNetv2 architecture.
The first is to adapt the depth of the network.
A MobileNetv2 architecture is made up of inverted bottleneck blocks~\cite{sandler2018mobilenetv2} that, over time, reduce the resolution of the input image while greatly increasing its channels (also often called features).
These blocks are divided into seven overall stages, see \cref{fig:mobilenet-figure}, inside which the internal representation shape between the blocks is kept constant.
Because of this structure, it is possible to remove blocks from these stages as long as one block in each stage remains to update the internal representation shape for the next stage.
We keep the initial two MobileNetV2 blocks intact to enable them to extract low-level features for later layers.
With that in mind, we define our model depth search space as described in \cref{eq:model-depth-search-space}.

\begin{equation}
\begin{aligned}
    (b_{3}, b_{4}, b_{5}, b_{6}, b_{7}) &\in B_{3} \times B_{4} \times B_{5} \times B_{6} \times B_{7}\\
    \text{where} \quad
    B_{3} &= \{0,1,2,3\}\\
    B_{4} &= \{0,1,2,3,4\}\\
    B_{5} &= \{0,1,2,3\}\\
    B_{6} &= \{0,1,2,3\}\\
    B_{7} &= \{0,1\}
\end{aligned}
\label{eq:model-depth-search-space}
\end{equation}

We also enable the possibility of completely removing stages if downstream stages are also removed.
For example, it is possible to remove stage 6, given that stage 7 is also removed by attaching the final classification head to the end of stage 5.
Our search strategy ensures that infeasible architectures containing blocks after a completely removed stage are not generated during the search and that this does not adversely affect the probability of finding deeper architectures.
Adapting the depth of the network has no implications on the RAM requirements to store data but can, depending on the stage and block, have a large impact on the RAM consumption of internal model representations and persistent memory taken up by the total number of parameters in the model.

The second approach to adapting the MobileNetv2 architecture is to change the width of the network, that is, the number of channels created by the convolutional layers of the model.
The width of the network is controlled by a single alpha parameter $A = \{0.1,0.2,\cdots,1\}$.
For implementation reasons, this is achieved using masking layers as introduced in FBNetV2~\cite{wan2020fbnetv2}.
This change significantly impacts both the size of internal model representations and the total number of parameters in the model.

Due to the complexity of accurately modeling model latency and energy consumption accurately, our implementation limits its hardware-aware metrics to RAM memory and flash storage.
To calculate the total memory consumption of the TinyML systems, we assume a streaming data paradigm where only a single data sample is stored in memory at a time.
Our implementation supports storing the data or model in datatypes of various sizes; however, all experiments in this paper are conducted assuming 8-bit datatypes for storing both data, internal model representations, and model parameters.

A URL to an open-source repository containing the new Data Aware \gls{nas} implementation can be found in \ref{app:implementation_repository}.

\section{Results}
This section presents a comprehensive evaluation of the Data Aware \gls{nas} system described in \cref{sec:improved_dnas} through four complementary experiments.
All experiments are conducted on a Nvidia V100 system for 23 hours, with metrics such as Accuracy, Precision, and Recall all estimated on full \SI{32}{\bit} floating point models.

Before presenting the results of these four experiments, we first revisit results from prior work on Data Aware \gls{nas}~\cite{njor2023data} to present a self-contained and holistic evaluation of Data Aware \gls{nas}.
Second, we show experimental results that Data Aware \gls{nas} is able to find TinyML systems with better predictive performance than Hardware Aware \glspl{nas} (i.e., Non-Data Aware \gls{nas}) under the same resource constraints.
Third, we present results that show a Data Aware \gls{nas} outperforming a Hardware Aware \gls{nas} at any time constraint.
Fourth, we apply the Data Aware \gls{nas} across embedded systems of differing resource constraints, showing how Data Aware \gls{nas} can automatically adapt to resource constraints where Hardware Aware \gls{nas} necessitates manual engineering work.
Fifth and last, we compare the search speed of the current supernet-based Data Aware \gls{nas} to the previously published Data Aware \gls{nas} implementation, showcasing how the technical improvements highlighted in this paper allow for the discovery of more and better TinyML systems. 

Collectively, these experiments provide strong evidence that a Data Aware \gls{nas} produces superior TinyML models compared to traditional Hardware Aware \glspl{nas}, achieving better performance across any timescale.

\subsection{Resource Consumption Under Performance Constraints}
We first revisit the main results from the previous publication on Data Aware \gls{nas}~\cite{njor2023data}.
These results are not based on the Data Aware \gls{nas} algorithm described in \cref{sec:improved_dnas}, but on the simpler ToyADMOS dataset~\cite{koizumi2019toyadmos} and a more naive approach to Data Aware \gls{nas}.
This earlier method utilized evolutionary search over a layer-based search space.
For more details, we refer to the previous paper~\cite{njor2023data}.

\begin{figure}
    \centering
    \includegraphics[width=0.8\linewidth]{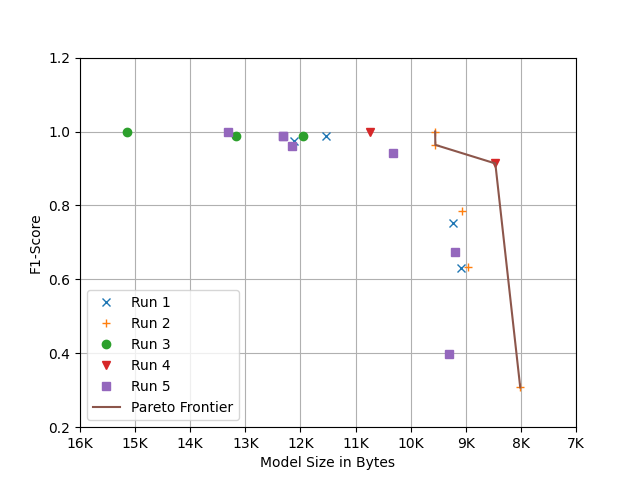}
    \caption{Pareto Frontier of TinyML Systems generated by the Data Aware \gls{nas} from~\cite{njor2023data}}
    \label{fig:experiment-1-data}
\end{figure}

\begin{figure}
    \centering
    \includegraphics[width=0.8\linewidth]{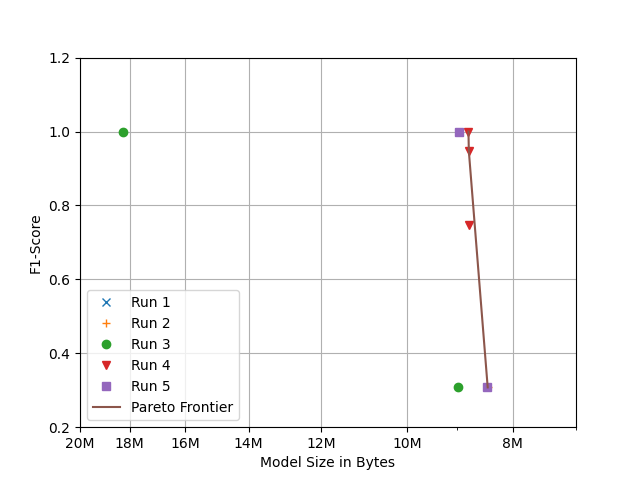}
    \caption{Pareto Frontier of TinyML Systems generated by the Hardware Aware \gls{nas} from~\cite{njor2023data}}
    \label{fig:experiment-1-fixed}
\end{figure}

In this prior work, the authors used both a Data Aware \gls{nas} and a traditional Hardware Aware \gls{nas} to search for TinyML systems that could effectively distinguish between the normal and anomalous samples of the ToyADMOS dataset.
We showcase the results of this experiment in Pareto Frontier plots in \cref{fig:experiment-1-data,fig:experiment-1-fixed}.
Note that while all points plotted represent Pareto optimal TinyML systems in each individual run, not all points are Pareto optimal across runs.
For that reason, we include a line drawing the Pareto Frontier of the Data Aware- and Hardware Aware \gls{nas} to each plot.

As evident from the TinyML systems reaching an F1-Score of 1 in these plots, both the Data Aware \gls{nas} and traditional Hardware Aware \gls{nas} are able to find models that can perfectly distinguish the normal and anomalous samples in the ToyADMOS test set.
However, the Data Aware \gls{nas} can do so with TinyML systems that consume more than four magnitudes less flash memory than the TinyML systems created by the traditional Hardware Aware \gls{nas}.
Unfortunately, the prior work did not track advanced resource consumption metrics such as RAM, but we expect it to have been significantly smaller on the systems created by Data Aware \gls{nas}.

These results provided experimental evidence that Data Aware \gls{nas} can be used to significantly reduce the resource consumption of TinyML Systems while maintaining performance.

\subsection{Predictive Performance Under Resource Constraints}\label{sec:results-predictive-performance}
The first experiment provides evidence that a Data Aware \gls{nas} can reduce the overall resource requirements of a TinyML system under constraints on the system's predictive performance.
A second central question is whether a Data Aware \gls{nas} can find better TinyML systems than a Hardware Aware \gls{nas} under identical resource constraints.

To investigate this, we set up an experiment where a Data Aware \gls{nas} and an equivalent Hardware Aware \gls{nas} search for a TinyML system to fit in memory constraints of \SI{512}{\kilo\byte} RAM and \SI{2}{\mega\byte} flash memory.
These memory constraints are typical for many TinyML-scale devices, for example, the STM32H743~\cite{lin2020mcunet} or the STMF767ZI~\cite{banbury2021micronets}.

\begin{figure}
    \centering
    \includegraphics[width=0.8\linewidth]{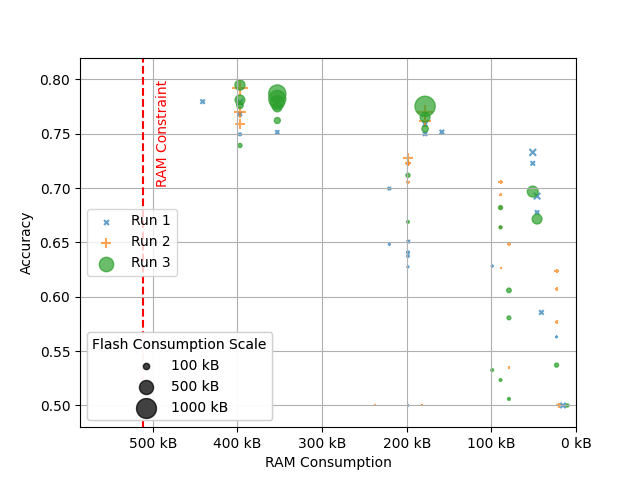}
    \caption{Pareto optimal TinyML systems generated using a Data Aware \gls{nas}}
    \label{fig:ex_2_data}
\end{figure}

\begin{figure}
    \centering
    \includegraphics[width=0.8\linewidth]{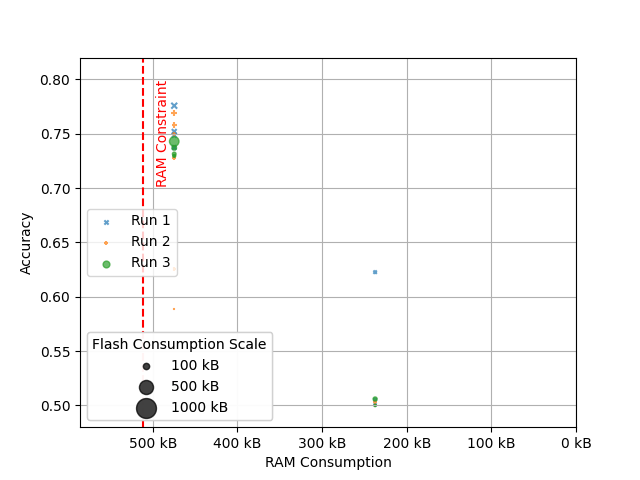}
    \caption{Pareto optimal TinyML systems generated using a traditional Hardware Aware \gls{nas}}
    \label{fig:ex_2_fixed}
\end{figure}

The results of this experiment are plotted in \cref{fig:ex_2_data,fig:ex_2_fixed} for the Data Aware- and Hardware Aware \gls{nas} respectively.
These figures show that the Data Aware \gls{nas} finds architectures that perform better than the Hardware Aware \gls{nas} at a maximum accuracy of $79.5\%$ and $77.6\%$, respectively.
Furthermore, the Data Aware \gls{nas} finds $22$ models with an accuracy of above $75\%$, whereas the Hardware Aware \gls{nas} finds only $4$ models above this threshold. 
The larger amount of Pareto optimal models generated by Data Aware \gls{nas} would, in a practical application, mean that it would be easier to find a TinyML system fitting the requirements for an application. 

These results suggest that a Data Aware \gls{nas} not only improves predictive performance but also provides more Pareto optimal choices, simplifying system selection for practical applications.

\subsection{Data Aware Neural Architecture Search Speed}\label{sec:results-search-time}
The search space of a Data Aware \gls{nas} is a superset of the search space of a Hardware Aware \gls{nas} and is several times larger.
Therefore, a relevant question is: Can a Data Aware \gls{nas} find better TinyML systems than a Hardware Aware \gls{nas} in a certain period of time?

\begin{figure}
    \centering
    \includegraphics[width=0.8\linewidth]{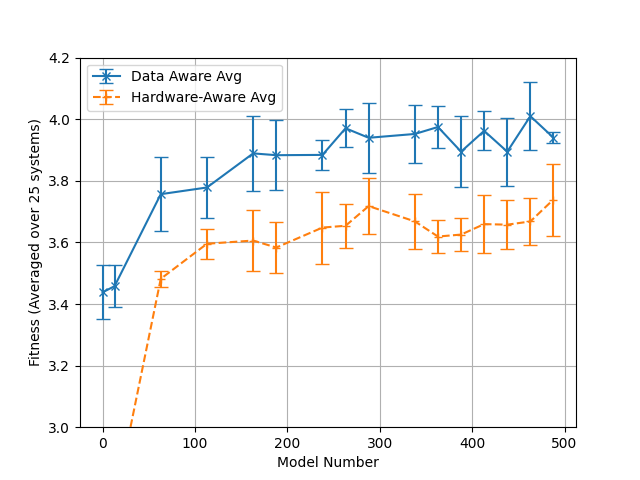}
    \caption{Fitness Evolution for both Data Aware and Hardware Aware \gls{nas}. (Fitness values averaged across 25 Models)}
    \label{fig:ex_3_evo}
\end{figure}

We use the models generated throughout the two experiments from \cref{sec:results-predictive-performance} to investigate this.
In \cref{fig:ex_3_evo}, we plot the fitness score yielded by the fitness function of our \glspl{nas} (see \ref{app:fitness_function_details}) for the models generated by both the Data Aware- and Hardware Aware \gls{nas}.

Contrary to our initial hypothesis, the models generated by the Data Aware \gls{nas} outperform the models generated by the Hardware Aware \gls{nas} throughout the entire search process, which suggests that it is worth using a Data Aware \gls{nas} over a Hardware Aware \gls{nas} regardless of the search time.

\begin{figure}
    \centering
    \includegraphics[width=\linewidth]{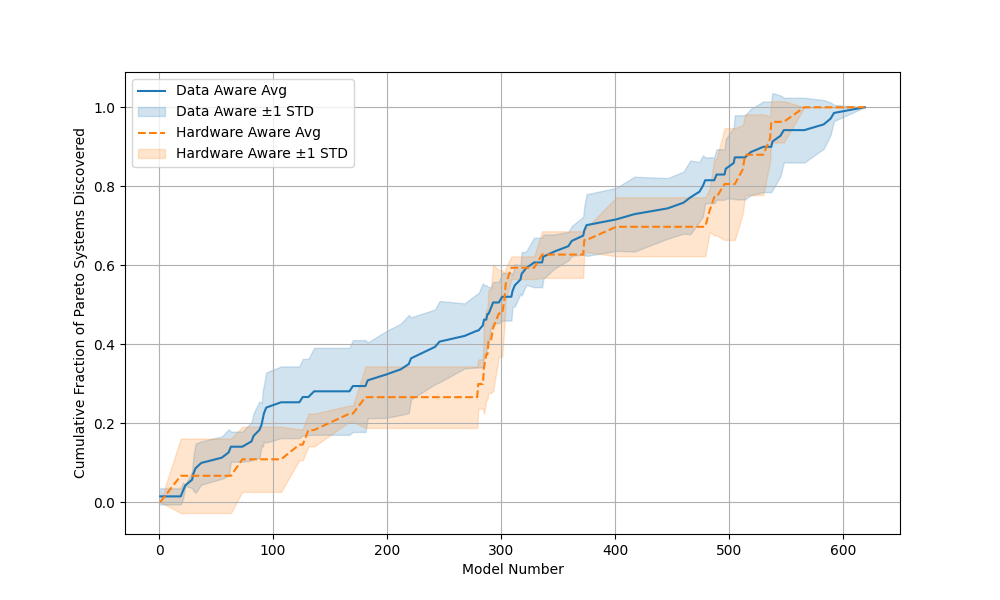}
    \caption{Cumulative fraction of Pareto optimal models discovered throughout Data Aware and Hardware Aware searches}
    \label{fig:ex_3_cumulative}
\end{figure}

Our fitness function is not a perfect proxy for when good TinyML systems are found, as systems can be Pareto optimal without achieving a high fitness score.
Therefore, as a complementary analysis, we plot the cumulative fraction of Pareto optimal discovered throughout both a Data Aware- and Hardware Aware \gls{nas} in \cref{fig:ex_3_cumulative}.

These results suggest that, while Pareto optimal systems are discovered throughout the entire search time in both \glspl{nas}, a Data Aware \gls{nas} will likely find more Pareto optimal models in short searches than traditional Hardware Aware \glspl{nas}.

\subsection{Ease of Use Across Resource Constraints}\label{sec:results-across-constraints}
The experiment in \cref{sec:results-predictive-performance} shows that Data Aware \gls{nas} can create better TinyML systems than Hardware Aware \gls{nas} under the resource constraints of that experiment.
A question remains: How does this generalize to other resource constraints?

To test this, we set up a Data Aware \gls{nas} to search for TinyML systems that can fit in the memory constraints of two other resource-constrained hardware platforms.
For the first we select memory constraints of \SI{320}{\kilo\byte} and \SI{1}{\mega\byte} corresponding to the STM32F746~\cite{lin2020mcunet,banbury2021micronets}.
As the second we set memory constraints of \SI{256}{\kilo\byte} and \SI{1}{\mega\byte} corresponding to the STM32F412~\cite{lin2020mcunet}, STM32F446RE~\cite{banbury2021micronets} or the Arduino Nano 33 \gls{ble} Sense~\cite{arduino-nano}.
Together, these systems form examples of a large, medium, and small TinyML hardware platform as visualized in \cref{tab:tinyml-hardware-platforms}.

\begin{table}
    \centering
    \begin{tabular}{lccc}
        \toprule
         & RAM & Flash & Example\\
        \midrule 
        Large & \SI{512}{\kilo\byte} & \SI{2}{\mega\byte} & STM32F765\\
        Medium & \SI{320}{\kilo\byte} & \SI{1}{\mega\byte} & STM32F746\\
        Small & \SI{256}{\kilo\byte} & \SI{1}{\mega\byte} & Arduino Nano 33 BLE Sense\\
        \bottomrule
    \end{tabular}
    \caption{The three examples of TinyML memory constraints considered in this paper.}
    \label{tab:tinyml-hardware-platforms}
\end{table}

\begin{figure}
    \centering
    \includegraphics[width=0.8\linewidth]{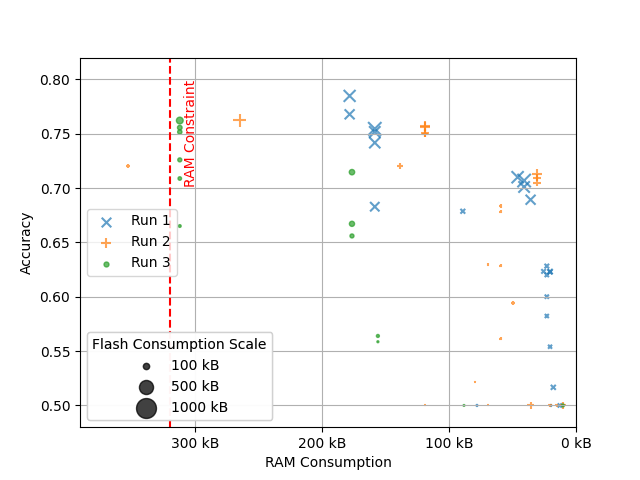}
    \caption{TinyML Systems Generated by Data Aware \gls{nas} for a device with constraints of \SI{320}{\kilo\byte} RAM and \SI{1}{\mega\byte} Flash Memory}
    \label{fig:ex_4_medium}
\end{figure}

\begin{figure}
    \centering
    \includegraphics[width=0.8\linewidth]{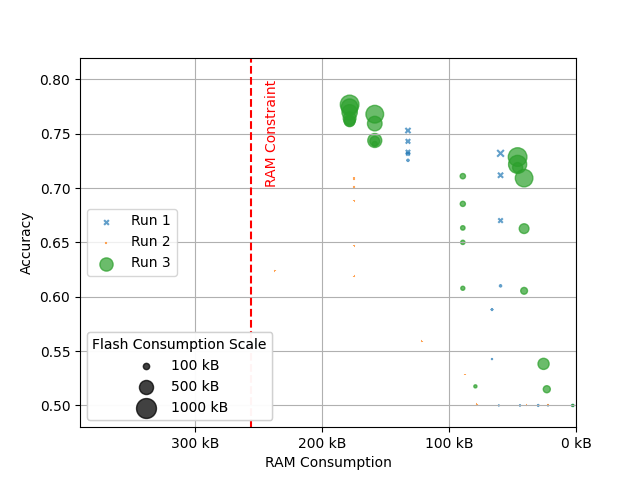}
    \caption{TinyML Systems Generated by Data Aware \gls{nas} for a device with constraints of \SI{256}{\kilo\byte} RAM and \SI{1}{\mega\byte} Flash Memory}
    \label{fig:ex_4_small}
\end{figure}

The TinyML Pareto frontier of TinyML systems generated for the large system can be found in \cref{fig:ex_2_data}, while the Pareto frontiers for the medium and small devices can be found in \cref{fig:ex_4_medium,fig:ex_4_small} respectively.

Notice that the Pareto optimal TinyML systems for the medium-sized device contain a system from run two that exceeds the RAM constraints of that device.
This is not a mistake but rather because not all constraint-violating models are excluded, only penalized by the fitness function (see \ref{app:fitness_function_details}) during our \gls{nas}.

These results show that our Data Aware \gls{nas} can discover suitable TinyML across different resource constraints with a maximum accuracy of $79.5\%$ in the large device, $78.5\%$ in the medium device, and $77.7\%$ for the small device.
Perhaps most importantly, these results are achieved by only changing the resource constraints given to the Data Aware \gls{nas}.
Thereby making a Data Aware \gls{nas} significantly easier to use than Hardware Aware \gls{nas}, which would have an engineer manually adjust many more parameters and/or data pre-processing to find TinyML systems across these resource constraints.

\subsection{Supernet Speedup}\label{sec:supernet_speedup}
In \cref{sec:improved_dnas}, we claim that it is necessary to introduce a new Data Aware \gls{nas} implementation based on MobileNetV2 supernets to find good TinyML systems for Person Detection in tractable time.
In this experiment, we seek to prove this by evaluating the supernet-based implementation's speedup over the original naive layer-based implementation proposed in the original work on Data Aware \gls{nas}~\cite{njor2023data}.

To do so, we run a modified version of the prior naive \gls{nas} implementation for 23 hours --- the same time as allocated to our new supernet-based \gls{nas} in other experiments. 
The modifications made to the naive \gls{nas} are all done in an attempt to enable a fair apples-to-apples comparison between the two \glspl{nas}.
A description of the modifications done can be found in \ref{app:non-supernet_configuration}.

Even with these modifications, creating a truly fair comparison is difficult due to the extent of changes made to improve the novel supernet-based search.
For example, the search space of the two different \glspl{nas} allows for discovering entirely different model architectures with varying training times and learning capacities.
Furthermore, the first models generated by the supernet-based search may take a long time to evaluate as the supernet for that data configuration needs to be pre-trained.
On the other hand, once the base supernets are trained, the supernet-based search should evaluate models much faster.

\begin{figure}
    \centering
    \includegraphics[width=0.75\linewidth]{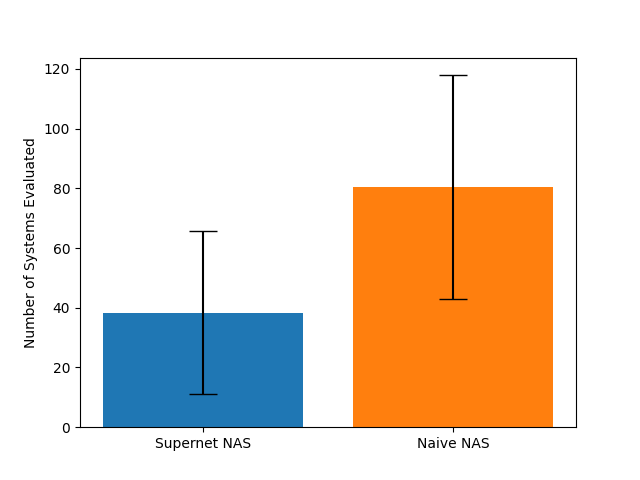}
    \caption{Average number of TinyML systems evaluated by the new supernet Based \gls{nas} and the prior naive layer based \gls{nas} from~\cite{njor2023data} before finding a system achieving over 67\% accuracy on the Wake Vision dataset. Standard deviation plotted in error bars.}
    \label{fig:ex-5}
\end{figure}

Because of these differences, a direct comparison between the number of TinyML systems generated during the 23 hours for each \gls{nas} would be unfair.
Instead, we compare the number of systems that must be evaluated before reaching a certain accuracy threshold.
Because the naive layer-based search space limits the accuracy that it is possible to find in 23 hours, we have to compare the two \gls{nas} at a low accuracy threshold at $67\%$ (The maximum non-decimal accuracy achieved by all naive \gls{nas} runs).
We plot the results of this experiment in \cref{fig:ex-5}.

These results show that the supernet-based \gls{nas} on average finds models $67\%$ accuracy in roughly half the number of system evaluations as the earlier naive \gls{nas}.
Furthermore, the supernet-based \gls{nas} finds much better systems throughout its search, reaching a maximum accuracy of $79.5\%$. 
In contrast, the naive \gls{nas} never finds a system reaching an accuracy above $70\%$.

\section{Future Work}
The search for TinyML systems using Data Aware \gls{nas} has only just begun.
There are still plenty of opportunities for future research into, e.g., improving Data Aware \gls{nas}, using it to create new state-of-the-art TinyML Systems, or applying it to discover theoretical insights into TinyML applications.
This section identifies some of the most promising of these future research directions.

\paragraph{Expanded Hardware Metrics}
Resource constraints on TinyML hardware encompass more than just the memory and computational constraints considered in this and other works on \gls{nas} for TinyML~\cite{lin2020mcunet, banbury2021micronets, garavagno2024colabnas, gambella2022cnas}.
Inference time and energy consumption metrics are especially tough to quantify as they depend heavily on target hardware.
Both are highly relevant metrics in TinyML system design.
Unfortunately, these metrics are currently hard to simulate or estimate accurately without deploying TinyML systems onto real hardware --- which is challenging in TinyML \gls{nas} where models must be evaluated quickly and without manual intervention.
Therefore, a promising research avenue to explore is to create accurate estimator tools that can provide realistic metrics in little time and be integrated automatically into \gls{nas} tools.

\paragraph{Differentiable Data Aware Neural Architecture Search}
Many state-of-the-art \glspl{nas} have recently explored using differentiable search spaces and strategies to speed up their search with promising results~\cite{liu2018darts, cai2018proxylessnas}.
Expanding Data Aware \gls{nas} to use a differentiable search space and strategy could improve search speeds to discover even better TinyML systems.
Doing so would require designing the data configuration search space in a way that makes it possible to use differentiable search methods.

\paragraph{TinyML Domain Insights using Data Aware Neural Architecture Search}
This paper shows how Data Aware \gls{nas} finds better-performing TinyML Person Detection systems on severely resource-constrained devices by lowering resolution and color representation.
This suggests that a larger \gls{nn} architecture is more important for Person Detection systems than a higher data granularity. 

TinyML systems in other application domains may perform better by trading off \gls{nn} architecture size for a larger data granularity. 
Investigating these trade-offs for different TinyML application domains may give novel insights into these domains that can prove valuable to the broader community.

\paragraph{Creating State of the Art TinyML Models using Data Aware Neural Architecture Search}
Although the experiments in this paper show how Data Aware \gls{nas} can produce better TinyML systems than an equivalent Hardware Aware \gls{nas}, the search space is not carefully designed to find state-of-the-art TinyML models.

Applying Data Aware \gls{nas} alongside a more carefully designed search space may be able to produce models that can outperform current state-of-the-art models such as the MCUNet models~\cite{lin2020mcunet}.

\section{Conclusion}
The adoption of TinyML in the industry remains slow, largely due to the deep expertise required to deploy heavily optimized TinyML systems to microcontroller scale devices.
AutoML approaches such as Hardware Aware \gls{nas} are promising avenues for making otherwise heavily technical optimization techniques accessible to the broader community.

In this paper, we extend prior work on Data Aware \gls{nas}, an augmentation of Hardware Aware \gls{nas}, which considers various ways to configure input data alongside a traditional Hardware Aware \gls{nas}.

As a part of this extension, we make significant technical improvements to prior work on Data Aware \gls{nas}, moving from considering trivial datasets, search strategies, and search spaces to a state-of-the-art TinyML dataset along with a supernet-based \gls{nas} utilizing a MobileNetV2 backbone.

On the basis of these improvements, we are able to provide new results showing Data Aware \gls{nas} discovering more accurate TinyML systems than Hardware Aware \gls{nas} given some resource constraints, doing so across any allocated \gls{nas} time and different resource constraints.

These results pave the way for using Data Aware \gls{nas} to design state-of-the-art TinyML systems and to gather new domain insights in TinyML that can be useful to all TinyML practitioners.

\section{Acknowledgements}\label{sec:acknowledgements}
This work was jointly supported by the Digital Research Centre Denmark (DIREC) and the Distributed Artificial Intelligent Systems (DAIS) project.
DIREC (https://direc.dk/) has received funding from the Innovation Fund Denmark under Grant 9142-00001B.
DAIS (https://dais-project.eu/) has received funding from the ECSEL Joint Undertaking (JU) under Grant 101007273. 
The JU was supported in part by the European Union’s Horizon 2020 research and innovation programme and Sweden, Spain, Portugal, Belgium, Germany, Slovenia, Czech Republic, Netherlands, Denmark, Norway, and Turkey.
Danish participants were supported in part by the Innovation Fund Denmark under Grant 0228-00004A.

\section{Declaration of generative AI and AI-assisted technologies in the writing process}
During the preparation of this work the author(s) used Grammarly and ChatGPT in order to improve the language of the paper. After using these tools/services, the authors reviewed and edited the content as needed and takes full responsibility for the content of the published article.

\appendix

\section{Implementation Repository}
\label{app:implementation_repository}
A repository containing the Data Aware \gls{nas} implementation used in this paper can be found at: \href{https://github.com/Ekhao/DataAwareNeuralArchitectureSearch}{\texttt{https://github.com/Ekhao/DataAwareNeural\linebreak{}ArchitectureSearch}}

\section{Fitness Function Details}
\label{app:fitness_function_details}
During both the Data Aware \gls{nas} and the traditional Hardware Aware \gls{nas} presented in this paper, TinyML systems are evaluated using a fitness function to determine how well they perform on both prediction and hardware-related metrics.
The fitness function used in this work can be found in \cref{eq:fitness_function}.

\begin{align}
f &= w_{1}a + w_{2}p + w_{3}r 
    + w_{4}\left( 1 - \frac{v_{r}}{x_{r}} \right)
    + w_{5}\left( 1 - \frac{v_{f}}{x_{f}} \right) \label{eq:fitness_function} \\
\text{where} \quad 
a &\quad \text{: Model accuracy,} \notag \\
p &\quad \text{: Model precision,} \notag \\
r &\quad \text{: Model recall,} \notag \\
v_{r} &\quad \text{: RAM violation (excess usage),} \notag \\
x_{r} &\quad \text{: Maximum allowable RAM consumption,} \notag \\
v_{f} &\quad \text{: Flash violation (excess usage),} \notag \\
x_{f} &\quad \text{: Maximum allowable Flash consumption,} \notag \\
w_{1}\cdots{}w_{5} &\quad \text{: Weights to adjust the importance of each term.} \notag
\end{align}

Maximum RAM and Flash consumption are configurable for each \gls{nas} run, and weights are equivalent, i.e., $w_{1}=w_{2}=\cdots{}=w_{5}$ in all of our experiments.

\section{Non-supernet Configuration}\label{app:non-supernet_configuration}
The Non-supernet Data Aware \gls{nas} which we utilize in \ref{sec:supernet_speedup}, is based on the state of the Data Aware \gls{nas} implementation at the \texttt{1f0cc3312\linebreak{}4d296b957dee1955a2106a72a419caa} commit hash in the repository described in \cref{app:implementation_repository} along with a few modifications.
The state of the repository at this commit hash can be found via the following URL: \href{https://github.com/Ekhao/DataAwareNeuralArchitectureSearch/tree/1f0cc33124d296b957dee1955a2106a72a419caa}{https://github.com/Ekhao/DataAwareNeuralArchitectureSearch/tree/\linebreak{}1f0cc33124d296b957dee1955a2106a72a419caa}

The following modifications have been made to the state of the repository at this commit hash for the experiment described in \cref{sec:supernet_speedup}:
\begin{itemize}
    \item The steps per epoch parameter for training the \gls{nn} model has been increased to 1000 steps.
    \item The number of training epochs has been increased to 25.
    \item Changed the \gls{nas} to run for 23 hours rather than a fixed number of models.
    \item The data search space has been updated to that used in this paper.
\end{itemize}

The changes introduced to the training steps are done in an effort to align the training effort that goes into training both supernet- and non-supernet-based models.
With these changes, a supernet-based model will be trained for 30,000 pre-training steps and 100 fine-tuning steps, whereas a non-supernet-based model will be trained for a total of 25,000 steps.

 \bibliographystyle{elsarticle-num} 
 \bibliography{cas-refs}





\end{document}